\documentclass{article}
\usepackage{spconf,amsmath,graphicx}
\usepackage{epsfig}
\usepackage{epstopdf}
\usepackage{url}

\title{Improving Spatial Codification in Semantic Segmentation}
\name{Carles Ventura,
 Xavier Gir{\'o},
  Ver{\'o}nica Vilaplana,
Kevin McGuinness,
Ferran Marqu{\'e}s,
 Noel E. O'Connor\thanks{This work has been developed in the framework of the project BIGGRAPH-
TEC2013-43935-R, financed by the Spanish Ministerio de Econom{\'i}a y Competitividad and the European Regional
Development Fund (ERDF). Thanks also to FPU-2010 Research Fellowship Program of the Spanish Ministry of Education.}}

\address{
Universitat Polit{\`e}cnica de Catalunya (UPC), Barcelona, Catalonia/Spain\\
Insight Centre for Data Analytics, Dublin City University (DCU), Ireland
}

\begin{document}
\ninept
\maketitle
\begin{abstract}
This paper explores novel approaches for improving the spatial codification for the pooling of local descriptors 
to solve the semantic segmentation problem. We propose to partition the image into three regions for each object to be
described: Figure, Border and Ground. This partition aims at minimizing the influence of the image context on the object
description and vice versa by introducing an intermediate
zone around the object contour. Furthermore, we also propose a richer visual descriptor of the object by applying
a Spatial Pyramid over the Figure region. Two novel Spatial Pyramid configurations are explored:
Cartesian-based and crown-based Spatial Pyramids. We test these approaches
with state-of-the-art techniques and show that they
improve the Figure-Ground based pooling in the Pascal VOC 2011 and 2012 semantic
segmentation challenges. 
\end{abstract}

\section{Introduction}

The classic approach to label the regions of an image with the appropriate object class has
been commonly based on SIFT-like \cite{Lowe04} and HOG-like \cite{Dalal05} features, pooled within each region using
Bag-of-Features (BoF)
\cite{Arbelaez12, Carreira12Object, Russakovsky12} or, more recently, Second Order Pooling (O2P) techniques
\cite{Carreira12, Yadollahpour13}. In addition, approaches based on convolutional neural networks (CNN) have gained
popularity among the
scientific community thanks to the results achieved by works such as \cite{Girshick13}, \cite{Hariharan14} and
\cite{Chen14}.
However, CNNs need to be pre-trained on large databases such as ImageNet Classification (1.2
million annotated images). 
In this paper, we investigate an alternative approach where features are manually designed instead of automatically
learned, reducing the need for large data collections and costly processing effort. 

Specifically, we propose to improve the visual description by partitioning the image into three regions (Figure,
Border and Ground) inspired by the work reported by Uijlings et al in \cite{Uijlings12}. Multiple authors have
highlighted the importance of the spatial context around an object during its
recognition \cite{Dalal05, Harzallah09, Felzenszwalb10}. 
In our work, we prove the potential of the Figure-Border-Ground (F-B-G) spatial pooling, extending the work in
\cite{Uijlings12} to the case of real object candidates and including new features in the visual description.

\begin{figure}[t]
\centering
\includegraphics[width=0.3\columnwidth]{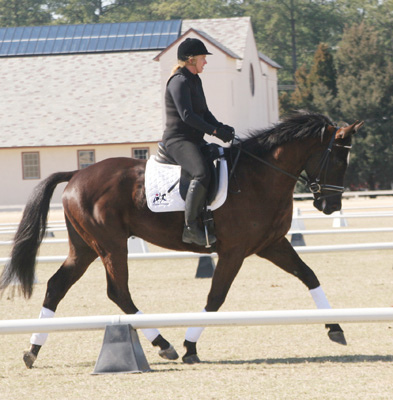}
\includegraphics[width=0.3\columnwidth]{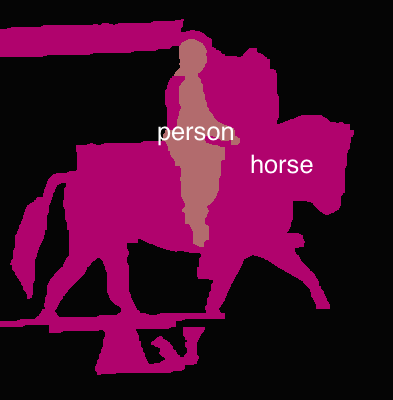}
\includegraphics[width=0.3\columnwidth]{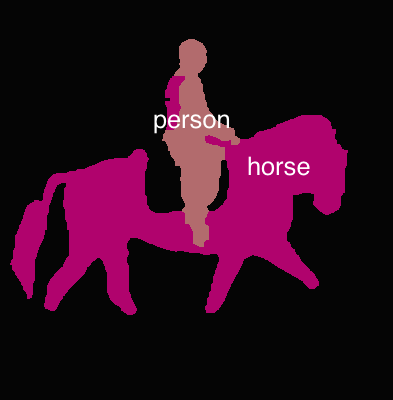}\\
\includegraphics[width=0.3\columnwidth]{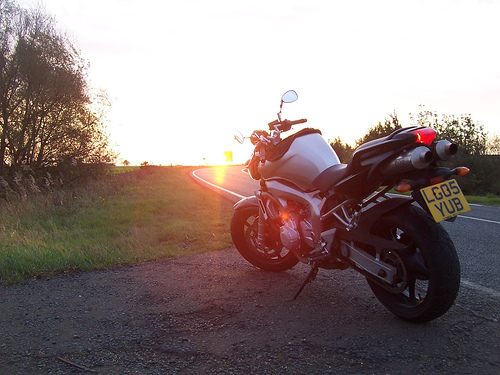}
\includegraphics[width=0.3\columnwidth]{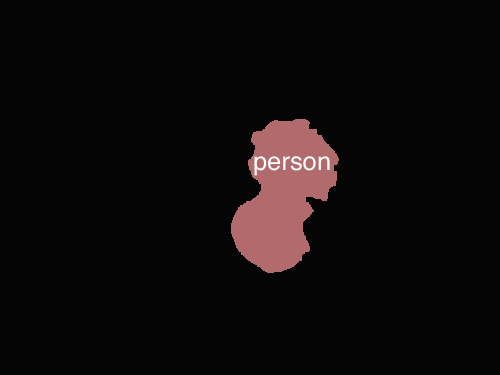}
\includegraphics[width=0.3\columnwidth]{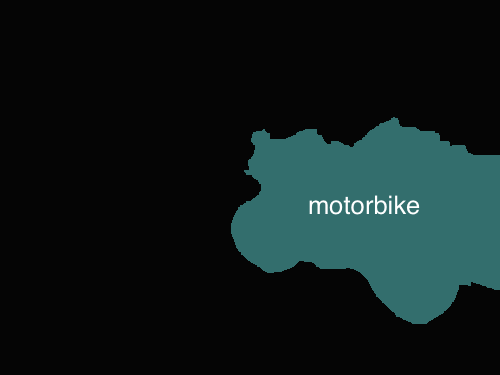}
\caption{Examples where a richer spatial codification improves the object segmentation and
recognition. Left: images to be semantic segmented. Middle: solution based on a Figure-Ground spatial pooling
\cite{Carreira12}. Right: solution based on a Figure-Border-Ground spatial pooling.}
\label{fig:attractive-solutions}
\end{figure}

On the one hand, our proposal has been tested over two
state-of-the-art object candidate algorithms: CPMC \cite{Carreira12cpmc} and MCG \cite{Arbelaez14}. Introducing the
Border pool for object candidates
represents a novel contribution with respect to the previous works \cite{Crandall07, Li10, Carreira12, Russakovsky12}
which only
considered Figure-Ground (F-G) spatial pooling. This intermediate area aims at minimizing the influence of
the image context in the object description and vice versa
as well as at capturing the rich contextual information located in the very neighbourhood
of the object itself.

On the other hand, our work also explores a novel approach for enriching the visual description of the object. We
propose to apply
a contour-based Spatial Pyramid (SP) over the Figure region using on two different configurations: $(i)$ a crown-based
SP, where the object is divided into different crowns for pooling, and $(ii)$ a Cartesian-based SP, where the object is
divided into four geometric quadrants for pooling.
These approaches for a richer spatial codification are combined with the O2P descriptors \cite{Carreira12}. Note that
both O2P and BoF solutions require significantly less training data than CNNs.



In the context of the Pascal VOC challenge named \emph{comp5}, the simplest training scenario implies only using
the annotations from the segmentation dataset, discarding the bounding box annotations from the detection dataset. In
that case, our approach improves the results from
\cite{Carreira12} with a performance gain of 12.9\%.
Figure~\ref{fig:attractive-solutions} shows two examples where the proposed richer spatial pooling based on a
F-B-G partition improves both the object segmentation and recognition with respect to a F-G
spatial pooling \cite{Carreira12}.



The remain of this paper is structured as follows. Section~\ref{sec:RW} gives an overview of the related work. In
Section~\ref{sec:contributions}, we present the main contributions of our work. Section~\ref{sec:experiments} gives
the experimental results. Finally, conclusions are drawn in Section~\ref{sec:conclusions}.


\section{Related Work}
\label{sec:RW}

Our work has been mainly inspired by \cite{Uijlings12}, where Uijlings et al investigated the impact of the visual
extent of an object
on the Pascal VOC dataset using a BoF with SIFT descriptors. Their analysis was performed in an ideal
situation where the ground truth object locations are used to create a separate representation with 3 types of regions:
the object's surrounding (Ground), near the object's contour (Border) and the object's interior (Figure). The authors in
\cite{Uijlings12}
reported a gain of 11.3\% in accuracy when introducing the Border.

The spatial coding of pooled features has not only been addressed from the perspective of taking automatically generated
regions as reference, but also through an arbitrary partition of the image. This is the case of the popular
Spatial Pyramid (SP) \cite{Lazebnik06}, which consists in dividing the whole image into a grid and pooling the
descriptors over each cell using a BoF framework. To our
best knowledge, the works \cite{Arbelaez12} and \cite{Gu12}, where a SP is applied over a bounding box
instead of at the image level, are the closest ones to our contour-based SP. There are also works such as
\cite{Chen12} where the layout of the SP depends on side information like object confidence maps or visual
salency maps, but it is also applied over the whole image.

To analyze our approaches for improving the spatial codification in semantic segmentation in a real context, we have
adopted a
solution based on the architecture proposed and released by Carreira et al in \cite{Carreira12}, which is briefly
described next. 150 CPMC object candidates \cite{Carreira12cpmc} are extracted per image and each object candidate is
described by its Figure and Ground features. Three types of enriched local
features (eSIFT, eMSIFT and eLBP) are densely
extracted and pooled using O2P \cite{Carreira12}.

\section{Contributions}
\label{sec:contributions}

Our proposal consists of two main contributions: $(i)$ the extension of the Figure-Border-Ground (F-B-G) pooling with
object candidates, and $(ii)$ a new contour-based Spatial Pyramid (SP) pooling to enrich the spatial information of
the object description.


\subsection{F-B-G pooling with object candidates}
\label{sec:FBG}

\begin{figure}
\centering
\includegraphics[width=0.32\columnwidth]{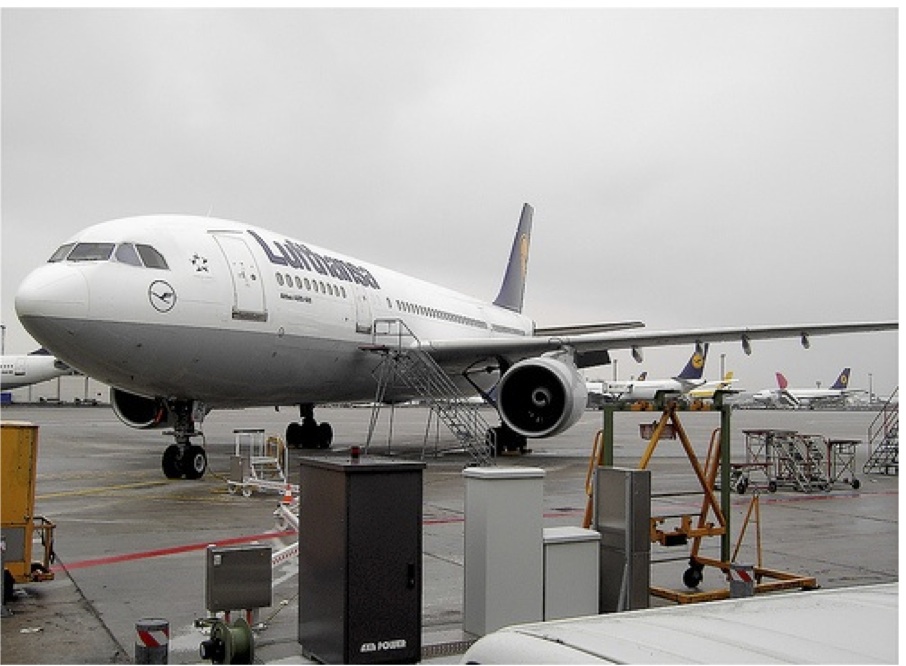}
\includegraphics[width=0.32\columnwidth]{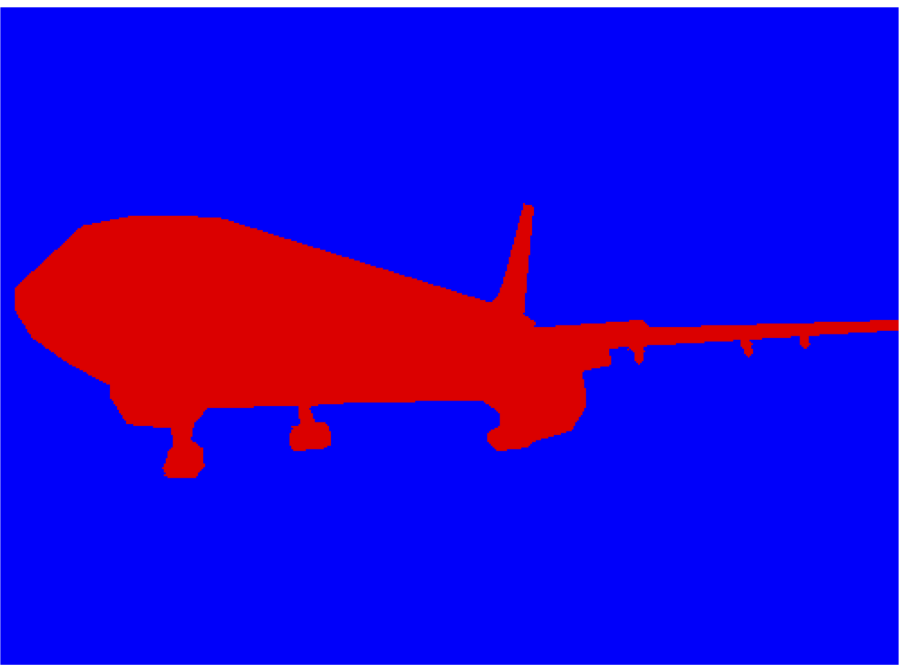}
\includegraphics[width=0.32\columnwidth]{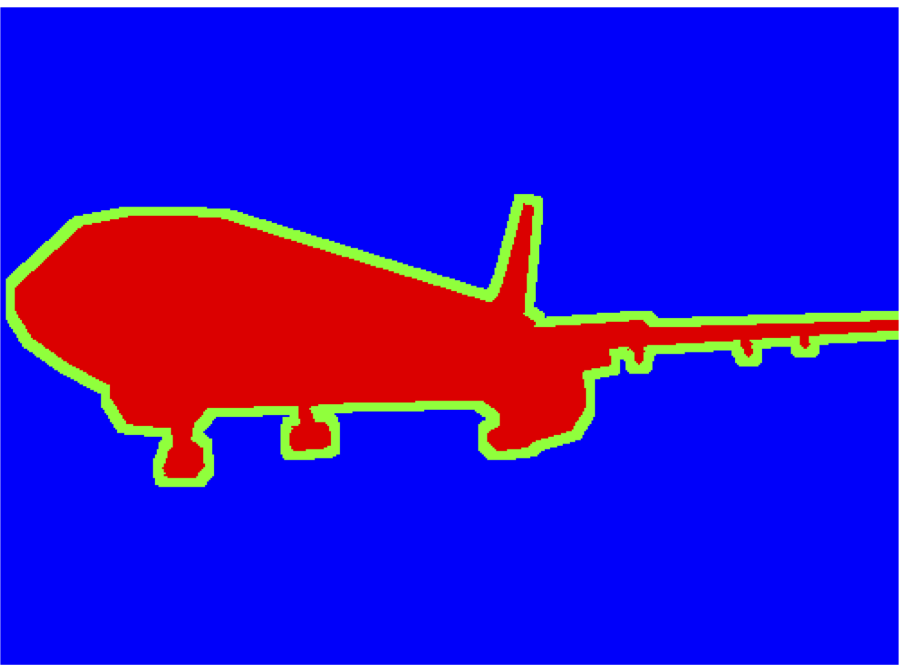}
\caption{Example of a Figure-Ground partition \cite{Carreira12} (in the middle) and a Figure-Border-Ground partition (on
the right) of the original image (on the left).}
\label{fig-figure-border-ground}
\end{figure}

In our work, we extend the spatial pooling based on a F-B-G image partition from \cite{Uijlings12}
by exploring its impact when applied in the realistic case of automatically extracted object candidates instead
of ground truth masks for the semantic segmentation challenge. As in \cite{Uijlings12}, we define the Border region as
a 5-pixel crown around the object. In contrast with \cite{Uijlings12}, we define a
region pool as the spatial layout where the local features can be centered independently of the extension of the
spatial support over which the local descriptors are computed. Therefore, the local descriptors extracted from a region
which are near the region contour can partially describe the neighbour region. In this way, we allow the use of
the usual 4$\times$4 SIFT descriptors as well as a multiscale dense feature detector instead of the 2$\times$2 SIFT
descriptors extracted at one single scale from \cite{Uijlings12}. Figure~\ref{fig-figure-border-ground} shows an
example of a F-G and a F-B-G image partitions.

This lack of absolute isolation of the description of each region pool can be justified in two ways. First,
multiple authors have highlighted the importance of the spatial context around an object during its
recognition \cite{Dalal05, Harzallah09, Felzenszwalb10}. Second, the fact that in our experiments, in
contrast with \cite{Uijlings12}, we also use a masked SIFT (MSIFT), which excludes any visual
information coming from the neighbour region. Therefore, the learning
process can automatically benefit from classes that can take advantage of the context (giving more importance to
non-masked descriptors) as well as from those where context can lead to confusion (giving more importance to masked
descriptors).


\subsection{Contour-based Spatial Pyramid}
\label{sec:SP}

\begin{figure}
\centering
\includegraphics[width=0.32\columnwidth]{figures/aeroplane.jpg}
\includegraphics[width=0.32\columnwidth]{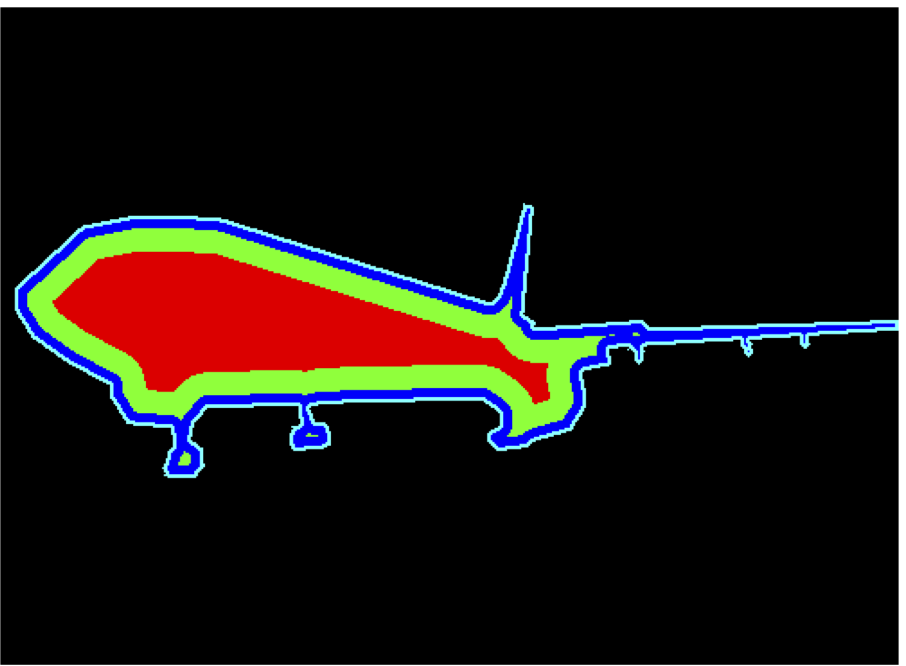}
\includegraphics[width=0.32\columnwidth]{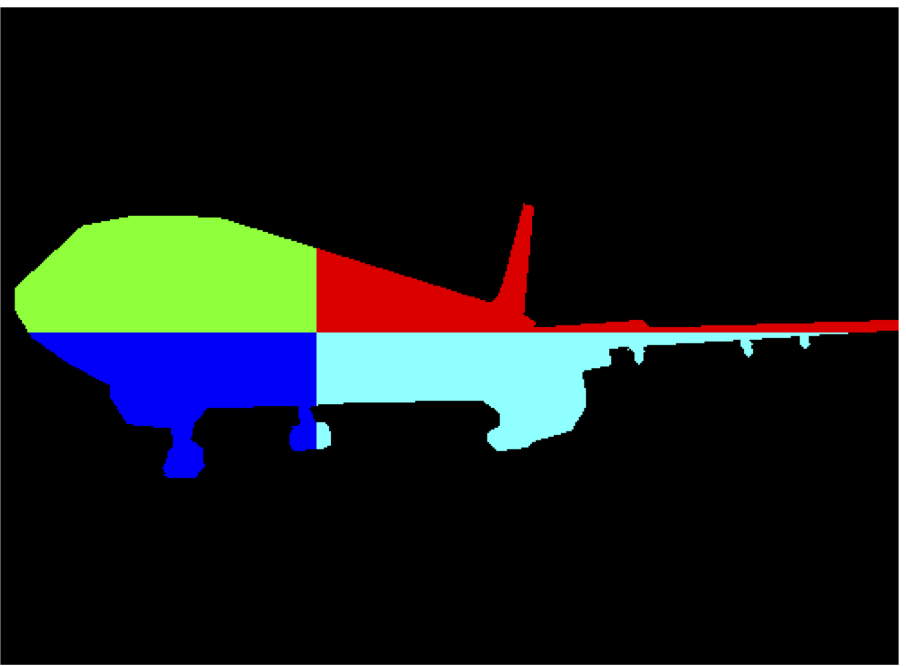}
\caption{Example of a 4-layer crown-based (in the middle) and a Cartesian-based (on the right) Spatial Pyramid from
an object mask of the original image (on the left).}
\label{fig-crown-sp}
\end{figure}

In a second contribution inspired by \cite{Lazebnik06}, we propose to apply a Spatial Pyramid (SP) coding approach over
the Figure
region to also improve the description of the interior of the object. More specifically, we apply a
SP centered on the object. We have performed an analysis based on two
different spatial configurations: $(i)$ a 4-layer crown-based SP, and $(ii)$ a Cartesian-based SP. The layers
of the crown-based SP are obtained by applying a
distance transform to the Figure mask. Then, the maximum value is used to define the different layers on a logarithmic
base. On the other hand, the
Cartesian-based SP divides the Figure region into 4 geometric quadrants which have the center of mass of the region as
origin. Figure~\ref{fig-crown-sp} shows an example of a 4-layer crown-based SP and a Cartesian-based SP.

\section{Experiments}
\label{sec:experiments}

The Pascal VOC Segmentation challenge \cite{Everingham10} provides a benchmark for semantic segmentation assessment. The
evaluation is performed by means of the Average of the Accuracy per Category (AAC), which is defined as the ratio
between the intersection and the union 
of the pixels classified as category $c_k$ and the pixels
annotated in the ground truth as $c_k$. The Pascal VOC Segmentation dataset is divided into three subsets: train,
validation and test. Preliminary experiments are performed using the train subset for training and the validation
subset for
test. Then, experiments are validated using both train and validation subsets for training and test subset for
test.



The experiments have been performed on the Pascal VOC 2011 and 2012 segmentation \emph{comp5} challenge, in which no
external data can be used for training. We address the realistic
scenario where a ranked list of pixel-wise object candidates are automatically generated. In our work, we have
considered the regions proposed by the CPMC \cite{Carreira12cpmc}, the same technique adopted in \cite{Carreira12},
since they allow a fair comparison of results. However, we have also considered the MCG \cite{Arbelaez14}, another
state-of-the-art technique for object candidate generation, to check the consistency of our contributions.


\subsection{Results with ideal object candidates}
\label{sec:GT-masks}
 
Experiments have been first performed using the ground truth object
masks (ideal object candidates). The use of these masks allows us to isolate pure recognition effects from segment
selection and
inference problems. This way it is possible to assess the improvements provided by the various spatial codifications in
an ideal scenario.



\subsubsection{F-B-G spatial pooling}

Table~\ref{Tab-GT} shows the results for different image spatial representations. The first and third columns
correspond to the configurations from \cite{Carreira12} where the Border region is included in the Ground
description. We propose two additional configurations: $(i)$ Figure(F)-Border(B), and
$(ii)$ Figure(F)-Border(B)-Ground(G).

On the one hand, the F-B configuration tries to answer the following question: How important is the entire
background in comparison with the bordering region? When eSIFT descriptors are pooled,
using only the Figure and Border regions and discarding the Ground is almost as
good as using the classical F-G partition of the whole image (66.24 and 66.43 respectively). If eMSIFT
descriptors are pooled instead, the average accuracy achieved by pooling them over F-B is even better than
over F-G (68.93 and 67.59 respectively). This indicates that the richest contextual
information for object recognition is located in the very near neighbourhood of the object itself.

On the other hand, the F-B-G configuration aims at showing the benefits of also including the rest of
the background as a region pool. Although pooling over Border can give better results than pooling over Ground
as seen before, Ground description still carries useful information for object recognition.

\begin{table}
\begin{center}
\begin{tabular}{|c||c|c|c|c|}
\hline
 & F \cite{Carreira12} & F-B & F-G \cite{Carreira12} & F-B-G \\
\hline\hline
eSIFT & 63.85 & 66.24 & 66.43 & \textbf{68.57} \\
\hline
eMSIFT & 64.81 & 68.93 & 67.59 & \textbf{70.84} \\
\hline
\end{tabular}
\end{center}
\caption{Gain of introducing the Border for pooling. Results using GT masks. Training over train11 and
evaluation over val11. F refers to Figure, B refers to Border and G refers to Ground.}
\label{Tab-GT}
\end{table}

Once eSIFT and eMSIFT have been independently analyzed, we explore the joint
combination of different descriptors by concatenation. This study
is performed to assess the impact of our proposal on the configuration with the best results obtained in
\cite{Carreira12}: with eSIFT-F, eSIFT-G, eMSIFT-F and eLBP-F (72.98).
Analogously, using only eSIFT and eMSIFT descriptors and the proposal of
partitioning the image into three regions (F-B-G) improves the average accuracy up to 73.84 (see
Table~\ref{Tab-SP-combination}) with
respect to the 72.48 obtained in \cite{Carreira12} (eSIFT and eMSIFT over F-G).

\subsubsection{Contour-based Spatial Pyramid}

In this section, we explore the proposal of improving the visual description by using the contour-based SP
presented
in Section~\ref{sec:contributions}. Table~\ref{Tab-SP} shows the results of applying the two Spatial Pyramids
configurations (crown-based and Cartesian-based) over the Figure region for the
eMSIFT descriptors.
The results show that both types of SPs give a
significative improvement of the average accuracy classification, especially when only the Figure region is
considered. Although the crown-based SP is better than the Cartesian-based SP for the Figure region, the
Cartesian-based SP gives the best performance when the Border and Ground regions are also considered. We believe that
this behavior is caused by the fact that the description of the Border region is more diverse with respect to the
geometric quadrants than the outermost layer of the crown-based SP.

\begin{table}
\begin{center}
\begin{tabular}{|c||c|c|c|c|}
\hline
 & F & F-B & F-B-G \\
\hline\hline
non SP & 64.81 \cite{Carreira12} & 68.93 & 70.84 \\
\hline
crown-based SP& \textbf{68.67} & 71.05 & 71.69 \\
\hline
Cartesian-based SP& 67.66 & \textbf{71.64} & \textbf{72.68} \\
\hline
\end{tabular}
\end{center}
\caption{Comparison between the non use of SP for the Figure region and the crown-based and
Cartesian-based SP approaches for GT masks. Training over train11 and evaluation over
val11.}
\label{Tab-SP}
\end{table}

The performance achieved by using only the eMSIFT descriptor (72.68) is almost as good as the accuracy achieved in
\cite{Carreira12} by combining eMSIFT, eSIFT and eLBP (72.98). 
Table~\ref{Tab-SP-combination} explores the joint
combination of different descriptors by concatenation when both Figure-Border-Ground spatial pooling and
Cartesian-based Spatial Pyramid are applied. As shown in this table, the use of both approaches improves the average
accuracy up to 75.86.

\begin{table}
\begin{center}
\begin{tabular}{|c|c|c|c||c|}
\hline
Figure & SP(F) & Border & Ground & AAC \\
\hline\hline
eS+eMS+eL & & & eS & 72.98 \cite{Carreira12}\\
\hline
eS+eMS & & eMS+eS & eMS+eS & 73.84 \\
\hline
eS+eMS+eL & eMS & eMS+eS & eMS+eS & \textbf{75.86} \\
\hline
\end{tabular}
\end{center}
\caption{Gain of introducing the Border for pooling, applying the Cartesian-based Spatial Pyramid over the
Figure (SP(F)) and combining eSIFT (eS),
eMSIFT (eMS) and eLBP (eL). Results using GT masks. Training over train11 and evaluation over val11.}
\label{Tab-SP-combination}
\end{table}

\subsection{Results with CPMC Object Candidates}

In this section, we
evaluate our two main contributions over CPMC object candidates. Note that there is a tight link between CPMC and the
O2P-based architecture from \cite{Carreira12} since these object candidates have
been reranked and filtered based on the same features used for classification, i.e. O2P features.

\subsubsection{F-B-G spatial pooling}
\label{sec:CPMC-FBG}

First, the experiments have been carried out in Pascal VOC 2011 using the train subset for training and the validation
subset for evaluation. 
The partitioning of the image for each object candidate into the Figure, Border and Ground regions improves the
performance up to 34.81 (with eSIFT) in comparison with the original partitioning into Figure and Ground regions (28.58
\cite{Carreira12}).


Next, we have performed experiments pooling the three different descriptors (eSIFT, eMSIFT
and eLBP) over the three proposed regions. The original performance achieved in
\cite{Carreira12} is 37.15. Our results from Table~\ref{Tab-CPMC-combination} show that using the partitioning of the
image into three regions for pooling the descriptors increases the average accuracy up to 38.91, which represents an
increase of 1.76 points.

\begin{table}
\begin{center}
\begin{tabular}{|c|c|c||c|}
\hline
Figure & Border & Ground & AAC \\
\hline\hline
eSIFT+eMSIFT+eLBP &  & eSIFT & 37.15 \cite{Carreira12} \\
\hline
eSIFT+eMSIFT+eLBP & eSIFT & eSIFT & \textbf{38.91} \\
\hline
\end{tabular}
\end{center}
\caption{Introducing the Border region with CPMC object
candidates. Training over train11 and evaluation over val11.}
\label{Tab-CPMC-combination}
\end{table}

For \emph{comp5}, the experiments have been carried out using only the segmentation annotations available for the train
and val sets of the segmentation challenge, discarding the bounding box annotations of the detection challenge. The
comparison between F-G and F-B-G poolings is shown in
Table~\ref{Tab-CPMC-comp5} for both Pascal VOC 2011 and 2012. The partitioning of the image into three regions
(F-B-G) gives the best performance, improving the average accuracy classification 5.0 and 2.3 points
with respect to the F-G pooling for VOC 2011 and VOC 2012 respectively. Notice that other results given by
the state-of-the-art techniques \cite{Li13, Xia12} have been obtained by using the bounding box annotations from the
detection challenge, which is out of the scope of this paper. 
Analyzing the results by categories, the F-B-G image partitioning improves the classification accuracy in
17 out of 20 categories in VOC 2011. In VOC 2012, 
the F-B-G approach improves the accuracy in 13 out of 20 categories.

\begin{table}
\begin{center}
\begin{tabular}{|c||c|c|}
\hline
 & F-G\cite{Carreira12} & F-B-G\\
\hline\hline
VOC11 & 38.8 & \textbf{43.8}\\
\hline
VOC12 & 39.9 & \textbf{42.2}\\
\hline
\end{tabular}
\end{center}
\caption{Results using CPMC object candidates for \emph{comp5} 2011 and 2012 and different image representations:
F-G and F-B-G}
\label{Tab-CPMC-comp5}
\end{table}

\subsubsection{Contour-based Spatial Pyramid}

Once the partitioning of the image into
three regions has been validated for CPMC object candidates, we proceed to validate the use of the Spatial Pyramid over
the Figure region. As before, the experiments are first evaluated over the validation subset. Using the Cartesian-based
SP over the Figure region with the eSIFT descriptor and ignoring both the Border and Ground regions
increases the perfomance up to 34.56, which is close to the improvement also achieved by the partitioning of the image
into three regions (34.81).

Applying both proposals, i.e. the Cartesian-based SP over the Figure region and the F-B-G
pooling, results in an average accuracy of 37.38. Notice that this result has
been achieved using only eSIFT, whereas the best perfomance achieved in \cite{Carreira12} is 37.15, which
uses a combination of eSIFT, eMSIFT and eLBP. An average accuracy of 39.62 is achieved when the three descriptors are
combined with the use of the three regions and the Cartesian-based SP
(see Table~\ref{Tab-CPMC-SP-combination}).


\begin{table}
\begin{center}
\begin{tabular}{|c|c|c|c||c|}
\hline
Figure & SP(F) & Border & Ground & AAC \\
\hline\hline
eS+eMS+eL & & & eS & 37.15 \cite{Carreira12} \\
\hline
eS & eS & eS & eS & 37.38 \\
\hline
eS+eMS & eS & eS & eS & 39.21 \\
\hline
eS+eMS+eL & eS & eS & eS & \textbf{39.62} \\
\hline
\end{tabular}
\end{center}
\caption{Results using CPMC object candidates for diferent image spatial representations and combining eSIFT (eS),
eMSIFT (eMS) and eLBP (eL) and applying the Cartesian-based Spatial Pyramid over Figure. Training over train11 and
evaluation over val11.}
\label{Tab-CPMC-SP-combination}
\end{table}


For \emph{comp5}, adding the Cartesian-based SP over the Figure region decreases the performance in 3.5
points for VOC 2011 (40.3) and 1.4 points for VOC 2012 (40.8). This decrease was not expected
based on the tendency shown in the previous experiments using the train set for training and the val set for evaluation
for both ground truth object masks and CPMC object candidates. The use of the SP over the Figure region
only improves the accuracy in 4 categories in VOC 2011 and in 8 categories in VOC 2012.

\subsection{Results with MCG Object Candidates}

Our spatial pooling approach has also been checked in another state-of-the-art object candidate generation:
Multiscale Combinatorial Grouping (MCG) \cite{Arbelaez14}. When the baseline solution given by
\cite{Carreira12} based on O2P features pooled over Figure-Ground is applied over MCGs instead of CPMCs, the average
accuracy drops to 30.88 with respect to the 37.15 achieved with CPMCs.

This drop in the performance seems to be in contradiction with the results reported in \cite{Arbelaez14} where for the
150 top-ranked
object candidates both techniques give a similar performance for segmentation (without considering recognition). We
believe that such a difference in the performance regarding the semantic segmentation is due to the fact that CPMCs have
been specifically reranked for the O2P-based architecture proposed in \cite{Carreira12}. Although about 800 CPMC generic
object candidates per image are extracted and ranked based on mid-level descriptors and Gestalt features, a linear
regressor also based on the O2P features is learned to rerank and filter them to generate the final pool of up to 150
CPMCs used in \cite{Carreira12}. Therefore, the features
used for classification (O2P) are also used for CPMC selection. On the other hand, MCG object candidates are ranked
based only on mid-level descriptors and Gestalt features.

However, we have also checked our spatial pooling proposals over the 150 top-ranked MCG object candidates. The
F-B-G spatial pooling increases the performance up to 34.09, which represents a gain of 3.21 points with
respect to the F-G spatial pooling (30.88). For such a spatial pooling, the classification accuracy is
improved for 15 out of 20 categories.

Furthermore, when the Cartesian-based SP is applied over the Figure region besides using the
F-B-G spatial pooling, the accuracy is increased up to 36.10, a gain of 2.01 points with
respect to the F-B-G pooling (34.09) and 5.22 points with respect to the F-G pooling (30.88).
Applying the Cartesian-based SP improves the accuracy for 16 out of 20 categories with respect to the
F-B-G pooling and for 19 out of 20 categories with respect to the original F-G pooling.

Although the results given by MCGs are worse than the ones achieved with CPMCs, we consider that these experiments
illustrate the robustness of our spatial pooling contributions with object candidates for semantic segmentation. 

\section{Conclusions}
\label{sec:conclusions}

We have presented two contributions for improving the spatial pooling beyond the classic Figure-Ground partitioning to
solve the semantic segmentation problem. 

On the one hand, we have extended the original idea from \cite{Uijlings12} where a Figure-Border-Ground spatial pooling
is applied in an ideal situation to a realistic scenario with the use of
object candidates. This richer spatial pooling has been tested with state-of-the-art
techniques (CPMC and MCG object candidates and O2P features), leading to improvements of the average accuracy in
all scenarios.

On the other hand, we have explored two different configurations (crown-based and Cartesian-based) of
Spatial Pyramid applied over the Figure region. Although this richer spatial pooling increased the performance when the
system was evaluated over the validation subset, this trend was not observed when it was
eventually assessed over the test subset.

Further visual results can be found in our project site\footnote{\url{https://imatge.upc.edu/web/publications/improving-spatial-codification-semantic-segmentation}}.
%

\bibliographystyle{IEEEbib}
\bibliography{icip_paper}

\end{document}